\pdfoutput=1
\documentclass[11pt]{article}

\usepackage{EMNLP2022}

\usepackage{times}
\usepackage{latexsym}

\usepackage{times}
\usepackage{latexsym}
\usepackage{graphicx} % DO NOT CHANGE THIS
\usepackage[ruled,vlined,noend]{algorithm2e}
\usepackage{enumitem}
\usepackage{booktabs}
\usepackage{float}

\usepackage[T1]{fontenc}
\usepackage[utf8]{inputenc}

\usepackage{microtype}

\usepackage{inconsolata}

\title{Progressive Sentiment Analysis for Code-Switched Text Data}

\author{
  Sudhanshu Ranjan $^1$ $\qquad$ Dheeraj Mekala $^1$ $\qquad$  Jingbo Shang $^{1,2}$ \\
  $^1$ University of California San Diego\\
  $^2$ Hal\i c\i o\u glu Data Science Institute, University of California San Diego\\
  \small \texttt{\{sranjan, dmekala, jshang\}@ucsd.edu}
}

\begin{document}
\maketitle
\begin{abstract}

Multilingual transformer language models have recently attracted much attention from researchers and are used in cross-lingual transfer learning for many NLP tasks such as text classification and named entity recognition.
However, similar methods for transfer learning from monolingual text to code-switched text have not been extensively explored mainly due to the following challenges:
(1) Code-switched corpus, unlike monolingual corpus, consists of more than one language and existing methods can't be applied efficiently,
(2) Code-switched corpus is usually made of resource-rich and low-resource languages and upon using multilingual pre-trained language models, the final model might bias towards resource-rich language. 
In this paper, we focus on code-switched sentiment analysis where we have a labelled resource-rich language dataset and unlabelled code-switched data. We propose a framework that takes the distinction between resource-rich and low-resource language into account.
Instead of training on the entire code-switched corpus at once, we create buckets based on the fraction of words in the resource-rich language and progressively train from resource-rich language dominated samples to low-resource language dominated samples. 
Extensive experiments across multiple language pairs demonstrate that progressive training helps low-resource language dominated samples.
% Our experiments verify that the progressive training helps low-resource language dominated samples.
% Extensive experiments across multiple language pairs demonstrate superior performance and significant advantages of our progressive training framework.
\end{abstract}

\section{Introduction}
% Para 1: What tasks and how is it being used? Where CS is important? Where CS is used?
% Para 2: Why can't these be used in the CS case? Talk about monolingual corpus which was used to train mBert.
% Para 3: How do you solve the problem? Refer to the figure for this

% Transfer learning in NLP has been explored rigorously over time (cite deepmoji, ulmfit, word embeddings). 
% transformer lang models -- how and where

% Learning text embeddings via pre-training on huge corpus, coupled with fine-tuning on sufficient supervised
% training data, have been ubiquitous recently in Natural Language Processing (NLP) and has shown success in a wide range of monolingual NLP tasks, mostly in English.
% More recently, BERT~\citep{devlin2019bert} was proposed that learns contextualized embeddings and achieves state-of-the-art results on several NLP tasks.
% Transformer models BERT ~\citep{devlin2019bert} \& GPT ~\citep{radford2018improving} learn the (bidirectional) contextualized embeddings for the words. For every task, the model is fine tuned separately. 
% Extending the idea, same model is trained for multiple languages simultaneously, thus making a single model learn multilingual representations. 
% Zero-shot learning for multilingual tasks have been explored by ~\citet{pires2019multilingual}. 

Code-switching is the phenomena where the speaker alternates between two or more languages in a conversation. 
% Multilingual speakers use code-switching extensively while communicating on social media. 
The lack of annotated data and diverse combinations of languages with which this phenomenon can be observed, makes it difficult to progress in NLP tasks on code-switched data.
And also, the prevalance of different languages is different, making annotations expensive and difficult.

% Recently, pretraining-finetuning paradigm is extensively used to reduce the need for annotated data~\citep{devlin2019bert, radford2018improving}.
% Learning text embeddings via pretraining on huge corpus, coupled with fine-tuning on small supervised
% training data, have been ubiquitous recently and has shown success in a wide range of monolingual NLP tasks, mostly in English.
% For e.g. BERT ~\citep{devlin2019bert} is a transformer-based language model that learns bidirectional contextualized embeddings and mBERT is the BERT model that is trained for multiple languages simultaneously, thus making a single model learn multilingual representations. 

\begin{figure}
    \centering
    \includegraphics[width=0.48\textwidth]{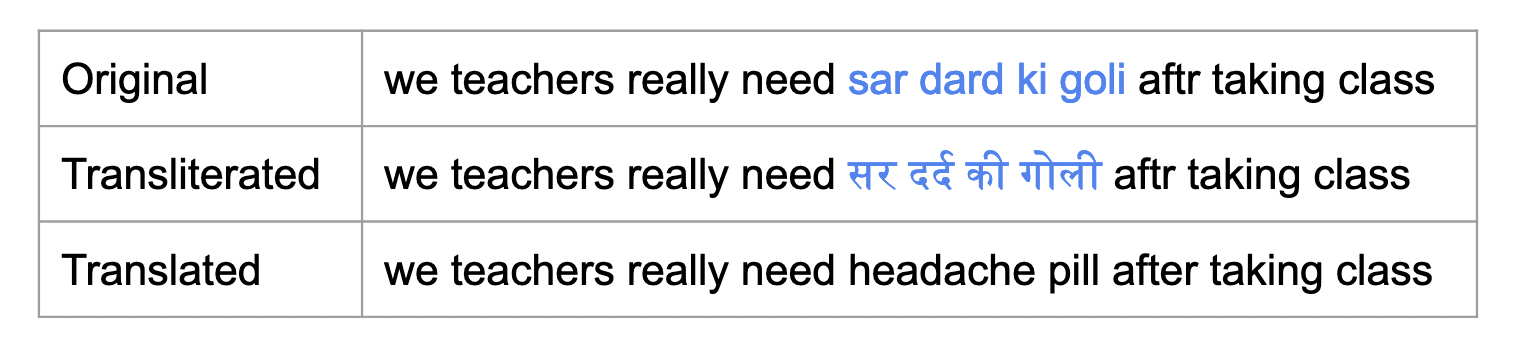}
    \caption{An example of code-switched text where words in both the languages together represent the sentiment. 
    % \sud{R1 comment}\dheeraj{A better example please if possible?}
    % , its transliterated and the translation versions. 
    A code-switched text generally contains phrases from multiple languages in a single sentence. The text in blue are words in Hindi that have been written in the Latin script.} 
    \label{fig1}
    \vspace*{-0.2cm}
\end{figure}

\begin{figure*}[t]
\centering
\includegraphics[width=0.99\textwidth]{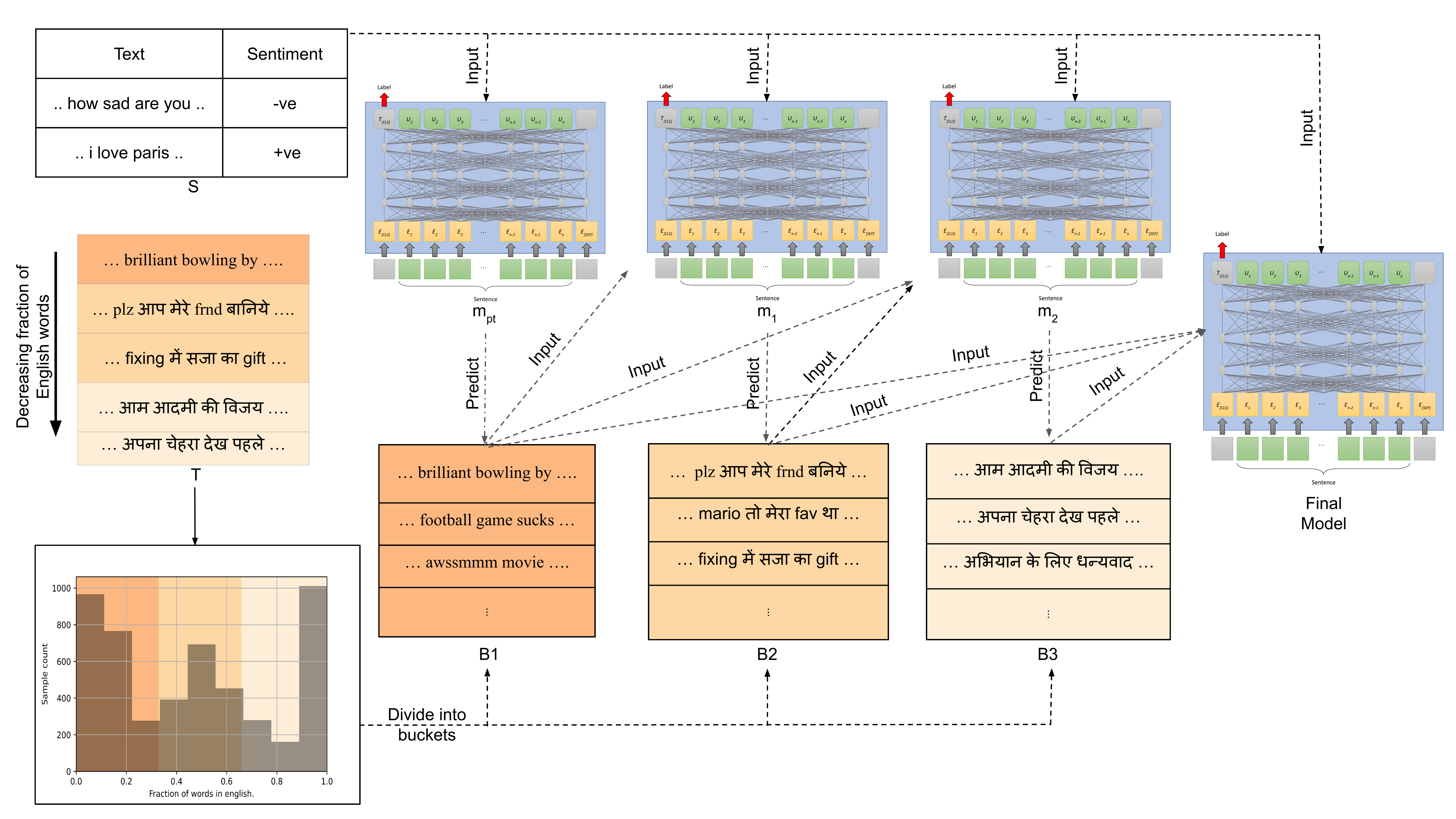} 
\vspace{-3mm}
\caption{A visualization of the progressive training strategy. The source labelled dataset $S$ in resource rich language should be easily available. Using $S$, a classifier is trained, say $m_{pt}$. Unlabelled code-switched dataset $T$ is divided into buckets using the fraction of English words as the metric. The leftmost bucket B1 has samples dominated by resource-rich language and as we move towards right, the samples in the buckets are dominated by low-resource language. $m_{pt}$ is used to generate pseudo-labels for unlabelled texts in bucket B1. We use texts from B1 along with their pseudo-labels and the dataset $S$ to train a second text classifier $m_1$. Then, $m_1$ is used to get the pseudo-labels for texts in bucket B2. We keep repeating this until we obtain the final model which is used for predictions.}
% \dheeraj{Rewrite this caption according to new figure.} Figure showing the progressive training framework. The target unlabelled dataset T is divided into buckets based on the fraction of English words in the sample. The initial model is trained using the labelled source dataset $S$. The pseudo-labels are generated for instances in resource-rich language dominated buckets. The source dataset and the high confidence instances from these buckets are then used to generate pseudo-labels for low-resource language dominated instances. Finally, high confidence samples from all buckets and source dataset $S$ are together used to train the final model which is then used to make final prediction.
% \dheeraj{(1.) Mention S and T in the figure. (2.) The flow should be Source dataset pretraining, bucket division, and progressive tarining. You didn't mention anything about bucket creation in the caption.}

\label{fig2}
\vspace{-3mm}
\end{figure*}

Intuitively, multilingual language models like mBERT~\citep{devlin2019bert} can be used for code-switched text since a single model learns multilingual representations. Although the idea seems straightforward, there are multiple issues.
Firstly, mBERT performs differently on different languages depending on their script, prevalence and predominance.  
mBERT performs well in medium-resource to high-resource languages, but is outperformed by non-contextual subword embeddings in a low-resource setting~\citep{heinzerling-strube-2019-sequence}.
% \dheeraj{cite the paper that mentioned this}. 
Moreover, the performance is highly dependent on the script~\citet{pires2019multilingual}.
% ~\citet{pires2019multilingual} showed that if the Hindi text is in Latin, the model performs worse compared to when it is Devanagari. 
% Even though mBERT has the ability to transfer between languages that are written in different scripts without being trained with a multilingual objective, zero-shot cross lingual ability across language pairs is not the same. Transfer learning between Urdu \& Hindi, English \& Bulgarian is more effective than English \& Japanese ~\citep{stickland2019bert}. 
Secondly, pre-trained language models have only seen monolingual sentences during the unsupervised pretraining, however code-switched text contains phrases from both the languages in a single sentence as shown in Figure~\ref{fig1}, thus making it an entirely new scenario for the language models. 
Thirdly, there is difference in the languages based on the amount of unsupervised corpus that is
% \dheeraj{shouldn't this "can be" be "is"} 
used during pretraining. For e.g., mBERT is trained on the wikipedia corpus. English has $\sim$ 6.3 million articles, whereas Hindi and Tamil have only $\sim$ 140K articles each. This may lead to under-representation of low-resource langauges in the final model. Further, English has been extensively studied by NLP community over the years, making the supervised data and tools more easily accessible. Thus, the model would be able to easily learn patterns present in the resource-rich language segments and motivating us to attempt transfer learning from English supervised datasets to code-switched datasets.
% \dheeraj{Remove the problems that are not addressed by our current framework.} 

The main idea behind our paper can be summarised as follows: \textit{When doing zero shot transfer learning from a resource-rich language (LangA) to code switched language (say LangA-LangB, where LangB is a low-resource language compared to LangA), the model is more likely to be wrong when the instances are dominated by LangB. Thus, instead of self-training on the entire corpus at once, we propose to progressively move from LangA-dominated instances to LangB-dominated instances while transfer learning}. 
% \dheeraj{Refer to figure 2 while explaining the framework below.}
As illustrated in Figure \ref{fig2}, model trained on the annotated resource-rich language dataset is used to generate pseudo-labels for code-switched data. Progressive training uses the resource-rich language dataset and (unlabelled) resource-rich language dominated code-switched samples together to  generate better quality pseudo-labels for (unlabelled) low-resource language dominated code-switched samples. Lastly, annotated resource-rich language dataset and pseudo-labelled code-switched data are then used together for training which increases the performance of the final model.  
% \sud{Please review}.

% transfer learning}. 
% % \dheeraj{Refer to figure 2 while explaining the framework below.}
% Model trained on the annotated resource-rich language dataset is used to generate pseudo-labels for code-switched data. 
% % Annotated resource-rich language dataset and pseudo-labelled code-switched data are then used together for the final training. 
% Progressive training uses the resource-rich language dataset and (unlabelled) resource-rich language dominated code-switched samples together to  generate better quality pseudo-labels for the (unlabelled) low-resource language dominated code-switched samples which increases the final performance of the model \dheeraj{The flow of this part is a bit confusing. Why did you discuss about final model training first and then high quality pseudo labels next?}
% \sud{final performance or model or performance of final model?}.

% \sud{Ack}
% \dheeraj{Replace langA with resource-rich after the italic content} \sud{Done}

Our key contributions are summarized as:
\begin{itemize}[nosep, leftmargin=*]
    \item We propose a simple, novel training strategy that demonstrates superior performance. 
    % Our strategy is simple enough to be implemented within a few lines of code.
    Since our hypothesis is based  on the pretraining phase of the multilingual language models, it can be combined with any transfer learning method.
    % \sud{Is this any TL method or multilingual model?}
    \item We conduct experiments across multiple language-pair datasets, showing efficiency of our proposed method.
    % \sud{We conduct extensive experiments across multiple language-pair datasets  }
    \item We create probing experiments that verify our hypothesis. 
    % \sud{I think this is your edit but grammar seems off, let me know if that's intended}
\end{itemize}
\noindent\textbf{Reproducibility.} Our code is publicly available on github \footnote{\url{https://github.com/s1998/progressiveTrainCodeSwitch}}.
% \footnote{\url{https://github.com/s1998/progressiveTrainCodeSwitch}}. 
% The main assumptions of our methodology are: (1) one of the languages is a high resource language (2) the model performance on high-resource  language dominated instances is better.\dheeraj{Why do we need to assume this?} \sud{You: If we don't assume  this, than the final performance would deteriorate.}

\vspace*{-0.2cm}
\section{Related work}
% We discuss related work from two angles.

% \textbf{Code-Switched NLP.} 
Multiple tasks like Language Identification, Named Entity Recognition, Part-of-Speech, Sentiment Analysis, Question Answering and NLI have been studied in the code-switched setting. For sentiment analysis, ~\citet{vilares-etal-2015-sentiment} showed that multilingual approaches can outperform pipelines of monolingual models on code-switched data. ~\citet{lal-etal-2019-de} use CNN based network for the same. ~\citet{winata-etal-2019-hierarchical} use hierarchical meta embeddings to combine multilingual word, character and sub-word embeddings for the NER task. \citet{aguilar-solorio-2020-english} augment morphological clues to language models and uses them for transfer learning from English to code-switched data with labels. ~\citet{samanta2019improved} uses translation API to create synthetic code-switched text from English datasets and use this for transfer learning from English to code-switched text without labels in the code-switched case. \citet{Qin2020CoSDAMLMC} use synthetically generated code-switched data to enhance zero-shot cross-lingual transfer learning. Recently, \citet{khanuja-etal-2020-gluecos} released the GLUECoS benchmark to study the performance of multiple models for code-switched tasks across two language pairs En-Es and En-Hi. The benchmark contains 6 tasks, 11 datasets and has 8 models for every task. 
% Multilingual transformers outperform the generic methods based on word and sentence embeddings. 
Multilingual transformers fine tuned with masked-language-model objective on code-switched data can outperform generic multilingual transformers. 
% \dheeraj{the above two statements look same. Remove one.} \sud{the first sentence says transformers > word embeddings . second says mutilingual transformers + code-switched MLM finetuning > multilingual transformers. The first seems redundant, removed.}
Results from ~\citet{khanuja-etal-2020-gluecos} show that sentiment analysis, question answering and NLI are significantly harder than tasks like NER, POS and LID.  In this work, we focus on the sentiment analysis task in the absence of labeled code-switched data using multilingual transformer models, while taking into account the distinction between resource-rich and low-resource languages.
Although our work seems related to curriculum learning, it is distinct from the existing work.
Most of the work in curriculum learning is in supervised setting~\cite{zhang-etal-2019-curriculum,xu-etal-2020-curriculum} and our work focuses on zero-shot setting, where no code-switched sample is annotated.
Note that, this is also different from semi-supervised setting because of distribution shifts between labeled resource-rich language data and target unlabeled code-switched data. 
% \sud{Should we use high-resource and low-resource or resource-rich and resource-poor instead? - It seems a very minor flaw though}
% \dheeraj{Both are fine. We need to be consistent.} \sud{Will make this chnage later if time permits then}

% \textbf{Curriculum learning.} \sud{Meta reviewer comment}  \citet{xu-etal-2020-curriculum} and \citet{zhang-etal-2019-curriculum} use curriculum learning for natural language understanding and machine translation in supervised setting. However, our work focuses on the zero-shot setting, where no code-switched sample is annotated at all and only the resource-rich language has annotated data. This problem setting is significantly different from supervised setting. Note that, this is also different from semi-supervised setting because of distribution shifts between labeled resource-rich language data and target unlabeled code-switched data. 

\section{Preliminaries}
Our problem is a sentiment analysis problem where we have a labelled resource-rich language dataset and unlabelled code-switched data. From here onwards, we refer the labelled resource-rich language dataset as the source dataset and the unlabelled code-switched dataset as target dataset. 
% \dheeraj{What do you mean by source data? Define this} 
Since code-switching often occurs in language pairs that include English, we refer to English as the resource-rich language. The source dataset, say $S$, is in English and has the text-label pairs $\{(x_{s_1}, y_{s_1}), (x_{s_2}, y_{s_2}), ... (x_{s_m}, y_{s_m})\}$ and the target dataset, say $T$, is in code-switched form and has texts $\{x_{cs_1}, x_{cs_2}, ... x_{cs_n}\}$, where $m$ is significantly greater than $n$. The objective is to learn a sentiment classifier to detect sentiment of code-switched data by leveraging labelled source dataset and unlabelled target dataset.
% \sud{leveraging labelled source dataset and unlabelled target dataset; remove domain? I don't think we use it anywhere else}

\section{Methodology}
Our methodology can be broken down into three main steps: (1) Source dataset pretraining, which uses the resource-rich language labelled source dataset $S$ for training a text classifier. This classifier is used to generate pseudo-labels for the target dataset $T$.  (2) Bucket creation, which divides the unlabelled data $T$ into buckets based on the fraction of words from resource-rich language. Some buckets would contain samples that are more resource-rich language dominated while others contain samples dominated by low-resource language. (3) Progressive training, where we initially train using $S$ and the samples dominated by resource-rich language and gradually include the low-resource language dominated instances while training. For rest of the paper, pretraining refers to step 1 and training refers to the training in step 3. And, we also use class ratio based instance selection to prevent the model getting biased towards majority label.

\subsection{Source Dataset Pretraining}
Resource-rich languages have abundant resources which includes labeled data. 
Intuitively, sentences in $T$ that are similar to positive sentiment sentences in $S$ would also be having positive sentiment (and same for the negative sentiment). 
Therefore, we can treat the predictions made on $T$ by multilingual model trained on $S$ as their respective pseudo-labels.
% This would create a noisy pseudo-labeled dataset $T$. \dheeraj{why does the noisy pseudo-labeled dataset and the unlabeled target dataset have same notation?} 
This would assign noisy pseudo-labels to unlabeled dataset $T$.
% The pretraining step is a text classification task. We use mBERT with a single layer \dheeraj{very vague, what do you mean by single layer?} as the classifier. We split the dataset $S$ into ratio of $80-20$ for training and development split. The development split is used for early stopping, and we use this checkpoint for generating pseudo-labels for the dataset $T$.
% \dheeraj{You can remove all these details about train-val-test and move it to experimental settings.}
The source dataset pretraining step is a text classification task. 
% \dheeraj{Don't use the term pretraining step as it confuses with the pretraining of language model? or mention it clearly what pretraining step refers to} \sud{The step name was decided long back , lets not modify that now unless necessary.  Also, see 2nd last line of preliminaries. I have rephrased sentence to make it less confusing and  replaced -pretraining- with  -source dataset pretraining- }
Let the model obtained after pretraining on dataset $S$ be called $m_{pt}$. This model is used to generate the initial pseudo-labels and to select the instances to be used for progressive training.

\subsection{Bucket Creation}
% \dheeraj{Make the subsections continuous, like a flow, as if you are explaining a story. End each subsection with a statement like the previous subsection can be ended as: "This model is further used in Bucket creation" and this subsection can be started like, "After pre-training on resource-rich data"}
% Since progressive training relies on using resource-rich language dominated samples and low-resource language dominated samples at different iterations, we divide the dataset $T$ into buckets based on fraction of words in resource-rich language.
Since progressive training aims to gradually progress from training on resource-rich language dominated samples to low-resource language dominated samples, we divide the dataset $T$ into buckets based on fraction of words in resource-rich language.
This creates buckets that have more resource-rich language dominated instances and also buckets that have more low-resource language dominated instances as well. In Figure \ref{fig2}, we can observe that the instances in the leftmost bucket are dominated by the English, whereas the instances in the rightmost bucket are dominated by Hindi.
% \sud{Refer to figure 2 here to convey the idea?} 
% In the initial iterations of step 3, we use samples from the former buckets and in later iterations\dheeraj{progressively accumulating the buckets}, we use samples from all the buckets \dheeraj{I think we can move the previous sentence to the next section} \sud{modify this part}. 
More specifically, we define:
\begin{center}
$f_{eng}(x_i) = \frac{n_{eng}(x_i)}{n_{words}(x_i)}$
\end{center}
where $n_{eng}(x_i)$ and $n\_{words}(x_i)$ denotes the number of English words and total number of words in the text $x_i$. Then, we sort the texts in dataset $T$ in decreasing order of $f_{eng}(x_i)$ and create $k$ buckets $(B_1, ..., B_k)$ with equal number of texts in each bucket.  Thus, bucket $B_1$ contains the instances mostly dominated by English language and as we move towards buckets with higher index, instances would be dominated by the low-resource language. 

\begin{algorithm}[t]
\caption{Pseudocode for our progressive training framework. }\label{alg:algorithm}
\SetAlgoLined
\small
  \textbf{Input}: Source dataset $S$, target dataset $T$\\
  \textbf{Parameter}: Selection fraction $\delta$, number of buckets $k$\\
  \textbf{Output}: Predictions on target dataset $T$\\
%   \tcp{Backbone model} 
  Model $m_{bb}$ $\epsilon$ $(mBERT, MuRIL, IndicBERT)$ \\
  Model $m_{pt}$ $\leftarrow$ $m_{bb}$ trained on dataset $S$ \\
  \tcp{Bucketing step}
  $T'$ $\leftarrow$ (($f_{eng}(x_i)$, $x_i$, $m_{pt}(x_i)$) for $x_i$ in $T$) \\
  $T'$ $\leftarrow$ $reverse\_sorted(T')$ \\
  $(B_1, B_2, ..., B_k)$ $\leftarrow$ divide $T'$ into $k$ equal buckets \\
  \tcp{Class ratio based instance selection} 
  $X_{st} = \phi$ \\
  \For{$\mathbf{i}=0,1$} {
  $X_{class_i}$ $\leftarrow$ Samples in $T'$ predicted to be in class $i$ using $m_{pt}$ \\
  $X_{st} = X_{st}$ $\cup$ ($\delta$ most confident samples in $X_{class_i}$) \\
  }
%   \dheeraj{Check here?}
%   $X_{class_q}$ $\leftarrow$ Samples in $T'$ predicted to be in class $q$ \\
%   $X_{st}$ $\leftarrow$ $\cup_{q=0}^{q=1}$ ($\delta$ most confident samples in $X_{class_q}$) \\
  \tcp{Progressive training step}
  Model $m_0 \leftarrow m_{pt}$ \\
  \For{$\mathbf{i}=1$ to $k$} {
    $X_{st_i}$ $\leftarrow$ $X_{st} \cap B_i$ \\
    $T_i$ $\leftarrow$ $\cup_{r=1}^{r=i}$($(x, m_{r-1}(x))$ for $x$ in $ X_{st_r}$) \\
    Model $m_i$ $\leftarrow$ $m_{bb}$ trained on $S \cup T_i$  
    % \jingbo{maybe say something more general than $mBERT$?}
  }
  \textbf{Return} $m_k(T)$  
%   \jingbo{shall we return all models instead of only the last?}
\end{algorithm}

\subsection{Progressive Training}
% As the model $m_{pt}$ is obtained by fine-tuning on a resource-rich language dataset $S$, it's more likely to perform better on resource-rich language dominated instances.
As the model $m_{pt}$ is obtained by fine-tuning on a resource-rich language dataset $S$, it is more likely to perform better on resource-rich language dominated instances.
% As our model is pretrained using $S$ which is in resource-rich language, it’s more likely to perform better on resource-rich language dominated instances. 
Therefore, we choose to start progressive training from resource-rich language dominated samples.
However, note that the pseudo-labels generated for dataset $T$ are noisy, thus we sample high confident resource-rich language dominated samples to obtain better quality pseudo-labels for the rest of the instances.
% Therefore, we first train on resource-rich language samples to obtain better quality pseudo-labels for the rest of the instances. 
% Hence, we first train using the resource-rich language samples to obtain better pseudo-labels for the rest of the instances. As the pseudo-labels generated for dataset $T$ are noisy, we only select a fraction of samples from $T$ for progressive training. 

Firstly, we use $m_{pt}$ to obtain all the high confidence samples from dataset $T$ to be used for progressive training and their respective pseudo-labels.  Among the samples to be used for progressive training, we select the samples from $B_1$ and use them along with $S$ to train a second classifier which is further used to generate pseudo-labels for the rest of the samples to be used for progressive training. Then we select samples from $B_2$ and use them along with samples from previous iterations (i.e. samples selected from $B_1$ and $S$) to get a third classifier. We continue this process until we reach the last bucket and use the model obtained at the last iteration to make the final predictions.

More formally, we use $m_{pt}$ to select the most confident $\delta$ fraction of samples from the dataset $T$, considering probability as the proxy for the confidence. Let $X_{st}$ denote the $\delta$ fraction of samples with the highest probability of the majority class to be used for progressive training.
% subset of samples to be used for the training.\dheeraj{Connect $X_{st}$ with the confident fraction. Otherwise it's confusing what $X_{st}$ means.} 
Let $X_{st_i} = X_{st} \cap B_{i}$, where $X_{st_i}$ is the subset of samples from bucket $B_i$ that would be used for the progressive training. To train across $k$ buckets, we use $k$ iterations. Let $m_j$ denote the model obtained after training for iteration $j$ and $m_0$ refers to model $m_{pt}$. Iteration $j$ is trained using texts $((\cup_{i=1}^{j}X_{st_i})\cup S)$. The true labels for texts in $S$ are available and for texts $X_{st_i}$, predictions obtained using model $m_{i-1}$ are considered as their respective labels. The model obtained at the last iteration i.e. $m_k$ is used for final predictions.

\subsection{Class ratio based instance selection}
% To address the class imbalance in datasets, we perform class ratio based instance selection while selecting samples for progressive training. 
% Specifically, instead of selecting $\delta$ fraction of samples from the entire dataset $T$, we select $\delta$ fraction of samples per class.
Datasets frequently have a significant amount of class imbalance.
Therefore, when selecting the samples for progressive training, we often end up selecting a very small amount or no samples from the minority class which leads to very poor performance. Hence, instead of selecting $\delta$ fraction of samples from the entire dataset $T$, we select $\delta$ fraction of samples per class. Specifically, let $X_+$ and $X_-$ denote the set of samples for which the pseudo-labels are positive and negative sentiment respectively. For progressive training, we choose $\delta$ fraction of most confident samples from $X_+$ and $\delta$ fraction of most confident samples from $X_-$.

The pseudo-code for algorithm is in Algorithm~\ref{alg:algorithm}.

\section{Experiments}
We describe the details relevant to the experiments in this section and also elaborate on the probing tasks.

% \begin{figure*}[!hb]
% \centering
% \includegraphics[width=0.95\textwidth]{LaTeX/images/f1_score.png}
% \caption{Model performance comparison across buckets. Bucketwise F1 score for datasets. $B_1$ contains
%  English dominated instances and $B_2$ contains low resource language dominated instances.
% Values are reported across 5 runs, points on the same vertical line are from the same run.}
% \label{modelPerfAcrossBuckets}
% \end{figure*}

\subsection{Datasets}
For source dataset pretraining, we use the English Twitter dataset from SemEval 2017 Task 4 ~\citep{rosenthal-etal-2017-semeval}. We use three code-switched datasets for our experiments : Hindi-English ~\citep{patra2018sentiment}, Spanish-English ~\citep{vilares-etal-2016-en}, and Tamil-English ~\citep{chakravarthi-etal-2021-lre}. Hindi-English, Spanish-English are collected from Twitter and the Tamil-English is collected from YouTube comments. The statistics of the dataset can be found in Table \ref{datasetStats}. Two out of the three datasets have a class imbalance, the maximum being in the case of Tamil-English where the positive class is $\sim$5x of the negative class. We upsample the minority class to create a balanced dataset. 
% Additional dataset details can be found in appendix \ref{newL}.

%  \sud{R2 comment on language identification and transliteration to be included in the main paper} 
Most of the sentences in the datasets are written in the Roman script. The words in Hindi and Tamil are converted into the Devanagari script. We use the processed dataset provided by \citet{khanuja-etal-2020-gluecos} for Hindi-English and Spanish-English datasets. The processed version of Hindi-English dataset has the Hindi words in Devanagari script. Since we deal with low-resource languages for which tools might not be well developed, we use heuristics to detect the words in English. For Spanish-English  and Tamil-English datasets, we use the Brown corpus from NLTK \footnote{https://www.nltk.org/} to detect the English words in the sentence. Words that are not present in the corpus are considered of another language. For the Tamil-English dataset, we use the AI4Bharat Transliteration python library \footnote{https://pypi.org/project/ai4bharat-transliteration/} to get the transliterations. 
 
 \begin{table}[!ht]
\small
    \centering
    \caption{Statistics of datasets. SemEval2017 is a supervised English dataset and the rest are code-switched datasets.}
    \label{datasetStats}
    \begin{tabular}{l ccc}
    \toprule
        \textbf{Dataset} & \textbf{Total} & \textbf{Positive} & \textbf{Negative} \\
        \midrule
        SemEval2017 & 27608 & 19799 & 7809\\
        \midrule
        Spanish-English & 914 & 489 & 425\\
        Hindi-English  & 6190 & 3589 & 2601\\
        % Ma-En FIRE & 2846 & 2246 & 600\\ \hline
        Tamil-English  & 10097 & 8484 & 1613\\
    \bottomrule
        \end{tabular}
\end{table}
\subsection{Model training}
In all the experiments we use multilingual-bert-base-cased (mBERT) classifier. The supervised English dataset has a 80-20 train-validation split. 
% For pretraining with supervised English dataset, we use 4 epochs and choose the best weights using the validation set. For training, we use 4 epochs in all the iterations and use the weights obtained after fourth epoch. For final evaluation, we use the model obtained from the last iteration. The batch size is 64, sequence length is 128 and learning rate is 5e-5. These hyperparameters are same for all the models used during pre-training and training. Every iteration takes approximately $\sim$1-2 seconds and $\sim$12 GB of memory on a GPU. 
Following ~\citet{wang-etal-2021-x}, $\delta$ is set to $0.5$. 
We observe that in most datasets, the number of spikes in the distribution plot of $f_{eng}(x_i)$ is either 1 or 2. For example, we observe there are only two spikes for the Hindi-English dataset in Figure \ref{distPlotImg} in Appendix \ref{bktStats}. Therefore, we set $k$=2.
% More details about the buckets be found in Table \ref{fracEng} and Table \ref{cntEng} in Appendix \ref{bktStats}.
We train the classifier in both pre-training and training phase for 4 epochs.
During pretraining with supervised English dataset, we choose the best weights using the validation set. 
While training, we use the weights obtained after fourth epoch. 
For final evaluation, we use the model obtained from the last iteration.
Additional details about the  hyper-parameters can be found in Appendix \ref{hyperP}. 
% \dheeraj{Don't write "in appendix". Make a subsection in appendix, add a label to it and link it here.}.
% \dheeraj{Mention about the ratio of code-switched in each bucket on 4 datasets. Add small table or something. You can put it in appendix and refer it here.} 
% Although, our method is generic enough for $k$ buckets, the target dataset size is too small to see any significant difference with higher number of buckets. 

In the rest of the paper, we refer to the model pretrained on the resource-rich language source dataset as model $m_{pt}$, the model trained on source dataset along with bucket $B_1$ as $m_1$, and the model trained on source dataset along with the buckets $B_1$ and $B_2$ as $m_2$. $m_2$ is used for final predictions. 

\subsection{Evaluation}
% As the datasets show a significant amount of skewness between the two classes, we choose to report micro, macro and weighted f1 score as done in ~\citet{mekala-shang-2020-contextualized}. 
As the datasets are significantly skewed between the two classes, we choose to report micro, macro and weighted f1 scores as done in ~\citet{mekala-shang-2020-contextualized}. For code-switched datasets, 
we use all the samples without labels during the self-training. The final score is obtained using the predictions made by model $m_2$ on all the samples and their true labels.
% \dheeraj{rewrite about development split, make things clear here.} \sud{done}
% Since the constraint on the code-switched case is the lack of enough dataset, we use sentences from both the train and development split (without labels) for progressive training and report the final score obtained on these splits together. 
For each dataset, we run the experiment with 5 seeds and report the mean and standard deviation.

\begin{center}
\begin{table*}[t]
\small 
  \centering
  \caption{Model performance on the Spanish-English, Hindi-English \& Tamil-English dataset using mBERT. For Tamil-English dataset, the \textbf{\textit{- Ratio}} method increases the F1 score of positive class (which is the majority class) by $\sim2\%$ but F1 score of negative class drops by $\sim9\%$. Thus, we observe a performance improvement in weighted F1 score and micro F1 score but a decrease in macro F1 score.}
  \label{tab1}
  \resizebox{\linewidth}{!}{
    \begin{tabular}{l ccc ccc ccc }
    \toprule
          \textbf{Methods} & \multicolumn{3}{c}{\textbf{Spanish-English}} & \multicolumn{3}{c}{\textbf{Hindi-English}} & \multicolumn{3}{c}{\textbf{Tamil-English}} \\
    \cmidrule(lr){2-4} \cmidrule(lr){5-7} \cmidrule(lr){8-10}
           & Macro-F1 & Micro-F1 & Weighted-F1 & Macro-F1 & Micro-F1 & Weighted-F1 & Macro-F1 & Micro-F1 & Weighted-F1\\
    \midrule
           Supervised  & $71.1$ & $71.5$ & $71.6$ & $74.7$ & $74.9$ &  $75.1$ & $62.7$ & $84.3$ & $82.8$ \\
    \midrule
          DEC &  $75.7 \pm 0.4$	 & $75.3 \pm 0.3$	 & $75.5 \pm 0.5$   & $66.0 \pm 0.5$ &	$67.9 \pm 0.7$ &	$67.1 \pm 0.4$ &  $51.7 \pm 0.9$ &	$77.8 \pm 0.8$ &	$75.5\pm 0.6$ \\
          No-PT &  $76.2 \pm 0.6$	 & $76.3 \pm 0.6$	 & $76.3 \pm 0.6$   & $67.1 \pm 0.8$ &	$69.2 \pm 0.3$ &	$68.4 \pm 0.6$ &  $53.3 \pm 0.4 $ &	$78.6 \pm 0.6 $ &	$76.7 \pm 0.2 $\\
          ZS & $76.1\pm 0.7$ &	$76.2 \pm 0.7$ &	$76.0 \pm 0.7$ &		$63.8 \pm 0.9$ &	$67.4 \pm 0.3$ &	$65.6 \pm 0.7$ &		$52.7 \pm 0.6$ &	$79.1 \pm 0.1$ &	$76.3 \pm 0.2$ \\
          Unsup-ST & $73.9 \pm 1.7 $ &	    $74.3 \pm 1.6 $ &	$74.1 \pm 1.7 $ &	$66.6 \pm 1.5 $ &	$66.7 \pm 1.4 $ &	$66.8 \pm 1.3 $ &	$49.8 \pm 2.1 $ &	$69.2 \pm 5.3 $ &	$69.8 \pm 8.4 $ \\ 
          Ours  & $\mathbf{77.4 \pm 0.8}$	 & $\mathbf{77.5 \pm 0.8}$	 & $\mathbf{77.5 \pm 0.8}$ & $\mathbf{67.8 \pm 0.8}$ &	$\mathbf{69.9 \pm 0.5}$ &	$\mathbf{69.1 \pm 0.7}$ &	
          $\mathbf{53.1 \pm 0.4}$ &	$80.5 \pm 0.2$ &	$77.5 \pm 0.1$   \\ 
    \midrule
          \hspace{2pt} - Source & $76.5 \pm 0.7$ &	$76.7 \pm 0.6$ &	$76.6 \pm 0.7$ & $67.2 \pm 0.5$ &	$68.7 \pm 0.4$ &	$68.3 \pm 0.4$ & $ 52.7 \pm 0.7 $ & $ 79.2 \pm 1.0 $ & $ 76.8 \pm 0.6 $ \\
          \hspace{2pt} - Ratio & $77.1 \pm 0.7 $ 	& $77.3 \pm 0.6 $ 	& $77.2 \pm 0.6 $ 	& $64.7 \pm 0.5 $ 	& $68.4 \pm 0.2 $ 	& $66.5 \pm 0.4 $ 	& $49.9 \pm 0.3 $ 	& $\mathbf{83.3 \pm 0.1 }$ 	& $\mathbf{77.7 \pm 0.1} $ \\ 
    \bottomrule
        \end{tabular}
    }
\end{table*}
\end{center}

\vspace*{-0.6cm}
\subsection{Baselines}
We consider four baselines as described below:
\begin{itemize}[leftmargin=*,nosep]
        \item Deep Embedding for Clustering (\textbf{DEC}) ~\citep{xie2016unsupervised} has been used in  WeSTClass ~\citep{meng18} for self-training using unlabeled documents after pretraining on generated pseudo documents. We adapt DEC similarly to our setting, by pretraining on $S$ and self-training using DEC objective only on $T$. 
        % \dheeraj{Is there any citation for this paper?} 
        % - there is not, I adapted this baseline to our use case
        % initially trains the model on the source dataset $S$, obtains pseudo-labels on unlabeled data in $T$ and further trains on all the samples from unlabelled data along with their pseudo-labels using the soft labeling objective  as done in LOTClass~\citep{meng2020text}.
        \item No Progressive Training (\textbf{No-PT})  initially trains the model on the source dataset $S$. As done in ~\citep{wang-etal-2021-x}, it selects $\delta$ fraction of the code-switched data with pseudo-labels and trains a classifier on selected samples and the source dataset $S$ without any progressive training.
        \item Unsupervised Self-Training (\textbf{Unsup-ST})~\cite{gupta-etal-2021-unsupervised} starts with a pretrained sentiment analysis model and then self-trains using code-switched dataset. We use the default version which doesn't require human annotations. We use the model $m_{pt}$ to initiate the self-training for fair comparison.
        \item Zero shot (\textbf{ZS}) ~\citep{mbertCrossLing} denotes the zero shot performance when the model is pretrained on monolingual resource-rich language dataset $S$. 
\end{itemize}
We also compare with two ablation versions of our method, denoted by \textbf{\textit{- Source}} and \textbf{\textit{- Ratio}}. \textbf{\textit{- Source}} uses only the code-switched dataset with its corresponding pseudo-labels without the source dataset $S$ for training. \textbf{\textit{- Ratio}} chooses the most confident samples for training without taking the class ratio into account. 

We also report the performance in the supervised setting, denoted by \textbf{\textit{Supervised}}. For each dataset, we train the model only on dataset $T$ but use true labels to do the same. This is the possible upper bound.

% \jingbo{Please define \textbf{Supervised} here too. Also highlight that it may even work worse and also why}

% We consider two baselines. For both the baselines, the model is initially trained on the source dataset S. Then we do the following: a) Train the final model on the source dataset $S$ along with the $\delta$ fraction of the code-switched data with pseudo-labels as done in X-CLASS ~\citep{wang-etal-2021-x}. For this baseline, we keep all the parameters the same as the actual method. We call this baseline as  Baseline-hl. b) Train on all the samples from unlabelled data along with their pseudo-labels using the DCE ~\citep{xie2016unsupervised} objective as done in LOTCLASS ~\citep{meng2020text}. We call this baseline as Baseline-dce. For this baseline, we keep the learning rate as 1e-5 without a scheduler.
% \sud{Add the third baselines}

% \jingbo{is it possible to add more compared methods?} \sud{I think it won't be feasible in the given time}

\subsection{Performance comparison}
The results for all the three datasets are reported in Table~\ref{tab1}. In almost all the cases, we observe a performance improvement using our method as compared to the baselines, maximum improvement being upto $\sim1.2\%$ in the case of Spanish-English. 
The comparison between ZS and our method shows the necessity of target code-switched dataset and the comparison between No-PT and our method shows that progressive training has a positive impact. In most cases, the final performance is within $\sim10\%$ of the supervised setting. We believe our improvements are significant since the baselines are close to the supervised model in terms of the performance and yet our progressive training strategy makes a significant improvement. We report the statistical significance test results between our method and other baselines in Table \ref{significanceTests} in Appendix \ref{statSigApp}. In all the cases, we observe the p-value to be less than 0.001. 
% \dheeraj{Discuss the statistical significance test results here.} 
The progressively trained model for Spanish-English does better than its corresponding supervised setting, outperforming it by $\sim6\%$. We hypothesize, this is because of having a large number of instances in the source dataset $S$, the progressively trained model has access to more information and successfully leveraged it to improve the performance on target code-switched dataset. 
The comparison between our method and its ablated version  \textbf{\textit{- Source}} demonstrates the importance of source dataset while training the classifier. We can note that our proposed method is efficiently transferring the relevant information from the source dataset to the code-switched dataset, thereby improving the performance. On comparing our method with \textbf{\textit{- Ratio}}, we observe that using class ratio based instance selection improves the performance in two out of three cases. For the Tamil-English dataset, we observe that the weighted \& micro F1 score are higher for \textbf{\textit{- Ratio}} method but the macro F1 score is poor. This is because the F1 score of the positive class increases by $\sim2\%$ but F1 score of negative class drops by $\sim9\%$ when using \textbf{\textit{- Ratio}} method instead of ours. Since the datatset is skewed in the favor of the positive class, this lead to a higher weighted and micro F1 score.
% \sud{To DJ - please re-review}

% \dheeraj{How about making a subsection for the below paragraph?} - done
\subsection{Performance comparison across buckets}
\begin{figure}[!h]
\centering
\includegraphics[width=0.45\textwidth]{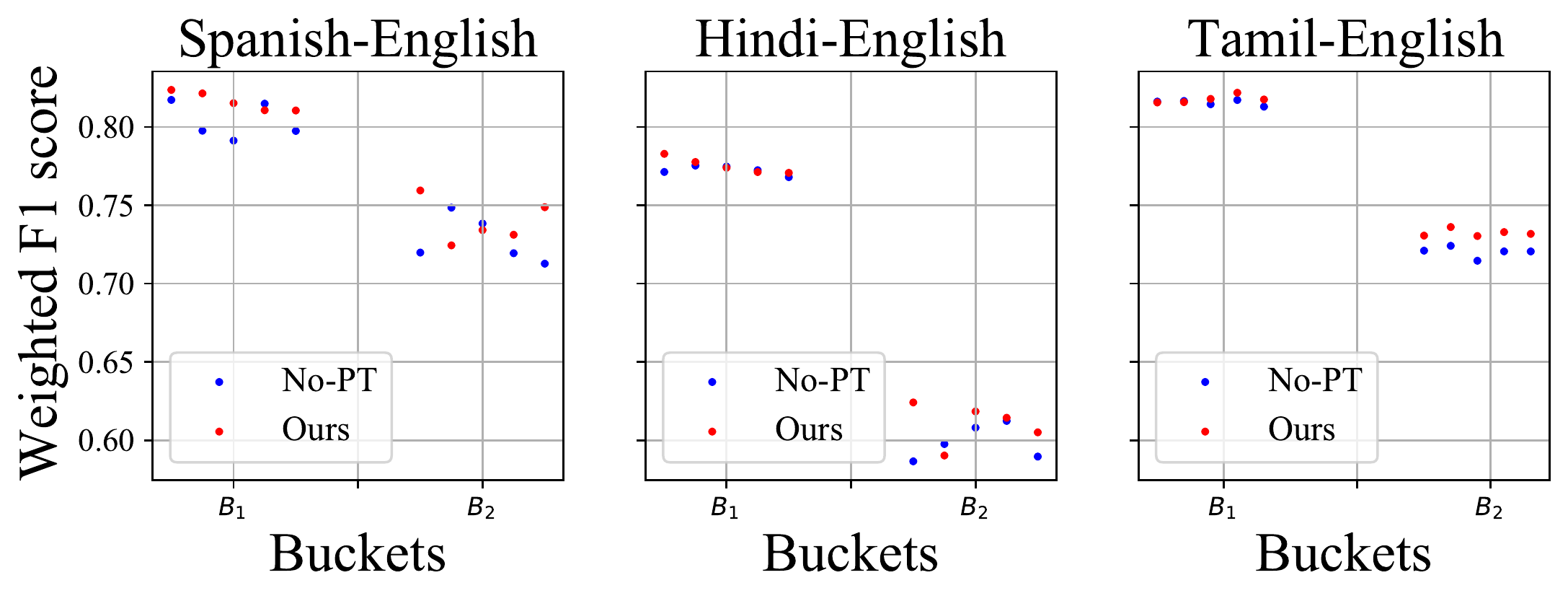}
\caption{Model performance comparison across buckets. $B_1$ contains English dominated instances and $B_2$ contains low resource language dominated instances. Values are reported across 5 runs. Points on the same vertical line are from the same run i.e. both No-PT and our model were initialized with same initial weights.}
\label{modelPerfAcrossBuckets}
\end{figure}

In  Figure~\ref{modelPerfAcrossBuckets}, we plot the performance obtained by No-PT and our method on both buckets. 
Since our method aims at improving the performance of low-resource language dominated instances, we expect our model $m_2$ to perform better on bucket $B_2$ and we observe the same.
As shown in Figure~\ref{modelPerfAcrossBuckets}, in most of the cases, our method performs better than the baseline on bucket $B_2$.
For bucket $B_1$, we observe a minor improvement in the case of Spanish-English, whereas it stays similar for other datasets. Detailed qualitative analysis is present in Appendix \ref{qualAna}.

\vspace*{-0.5cm}
\subsection{Probing task : Out-Of-Distribution (OOD) detection}
As previously mentioned, our proposed framework is based on two main hypotheses: (1.) A transformer model trained on resource-rich language dataset is more likely to be correct/robust on resource-rich language dominated samples compared to the low-resource language dominated samples, (2.) The models obtained using the progressive training framework is more likely to be correct/robust on the low-resource dominated samples compared to the models self-trained on the entire code-switched corpus at once.
To confirm our hypotheses, we perform a probing task where we compute the fraction of the samples that are OOD.
More specifically, we ask two questions: a) Is the fraction of OOD samples same for both the buckets for model $m_{pt}$? b) Is there a change in OOD fraction for bucket $B_2$ if we use model $m_1$ instead of model $m_{pt}$? 
The first question helps in verifying the first part of the hypothesis and the second question helps in verifying the second part of the hypothesis.

Since the source dataset $S$ is in English and the target dataset $T$ is code-switched, the entire dataset $T$ might be considered as out-of-distribution. However, transformer models are considered robust and can generalise well to OOD data ~\citep{hendrycks-etal-2020-pretrained}. Determining if a sample is OOD is difficult until we know more about the difference in the datasets. However, model probability can be used as a proxy. We use the method based on model's softmax probability output similar to ~\citet{hendrycks2018baseline} to do OOD detection. Higher the probability of the predicted class, more is the confidence of the model, thus less likely the sample is out of distribution. 

For a given model trained on a dataset, a threshold $p_{\alpha}$ is determined using the development set (or the unseen set of samples) to detect OOD samples. $p_{\alpha}$ is the probability value such that only $\alpha$ fraction of samples from the development set (or the unseen set of samples) have probability of the predicted class less than $p_{\alpha}$. For example, if $\alpha=10\%$, $90\%$ of samples in the development set have probability of predicted class greater than $p_{\alpha}$. If a new sample from another dataset (or bucket) has probability of predicted class less than $p_{\alpha}$, we would consider it to be OOD. Using $p_{\alpha}$, we can determine the fraction of samples from the new set that are OOD. Since, there is no method to know the exact value of $\alpha$ to be used, we report OOD using three values of $\alpha$ : {0.01, 0.05 and 0.10}. For model $m_{pt}$, we use the development split from the dataset $S$ to determine the value of $p_{\alpha}$, and for model $m_1$, we use the set of samples from bucket $B_1$ that are not used in self-training (i.e. $B_1-X_{st_1}$) to determine $p_{\alpha}$. Based on the value of $\alpha$, we conduct two experiments and answer our two questions.

%  The idea is to use a threshold $p_{\alpha}$ which depends on the samples of the test set. $p_{\alpha}$ is the probability value such that only $\alpha$ fraction of samples in the test set have probability of the predicted class less than $p_{\alpha}$. If a new sample while doing transfer learning has probability of the predicted class lesser than $p_{\alpha}$, we consider it OOD.  Using $p_{\alpha}$, we can determine how much of another dataset is OOD. Since, there is no method to know the exact value of $\alpha$ to be used, we report OOD using three values of $\alpha$ : {0.01, 0.05 and 0.10}. For model $m_{pt}$, we use the development split from the dataset $S$ to determine the value of $p_{\alpha}$, and for model $m_1$, we use the set of samples from bucket $B_1$ not used in self-training (i.e. $B_1-X_{st_1}$) to determine $p_{\alpha}$. Based on the value of $\alpha$, we conduct two experiments:\\

\begin{figure}[!h]
\centering
\includegraphics[width=0.48\textwidth]{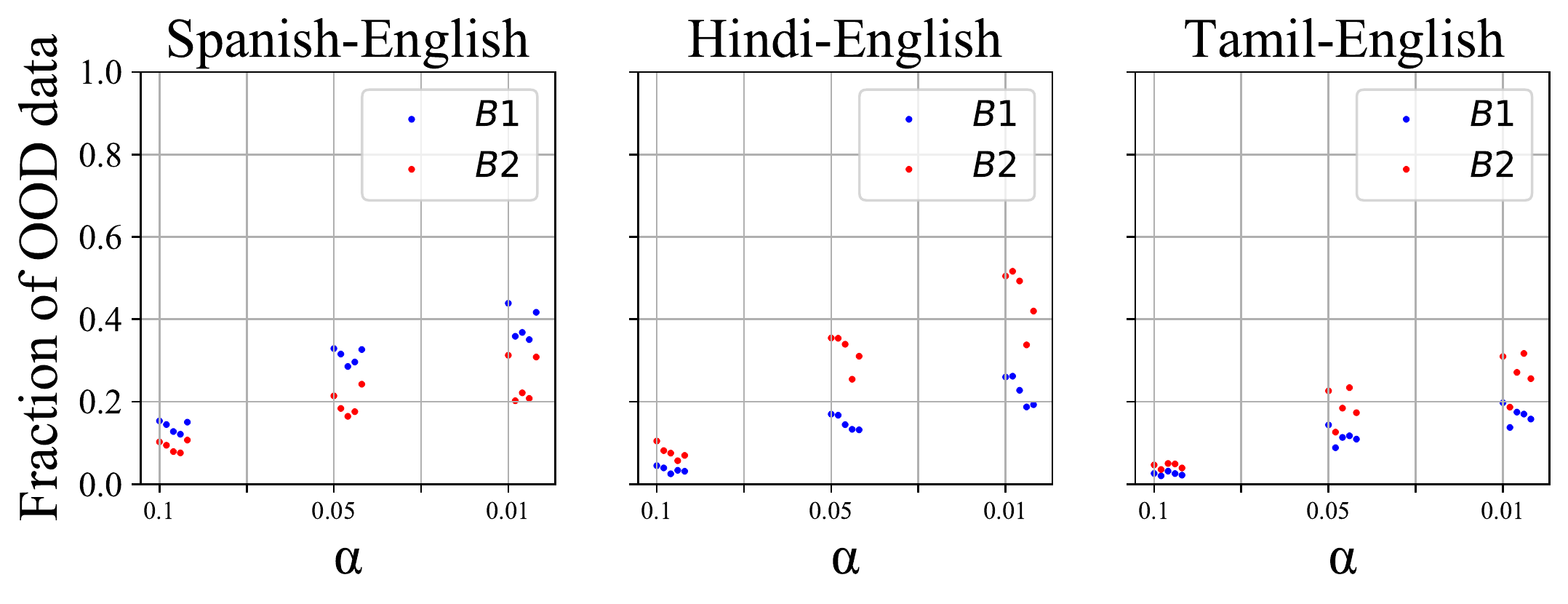} % Reduce the figure size so that it is slightly narrower than the column.
\caption{Figure showing bucketwise OOD for different datasets using the model $m_{pt}$. In most cases, samples in $B_2$ are more OOD compared to samples in $B_1$ across different values of $\alpha$. Values are reported across 5 runs, points on the same vertical line are from the same run i.e. once a model $m_{pt}$ has been trained, the same model is used to evaluate the fraction of OOD data in both the buckets.
% \dheeraj{Do you have time to change the color pairs here? They look very similar. May be choose like in Figure 3. Apply the same in figure 5}
% \sud{Do I need to explain the meaning of same run here? I expalined it once in figure 3.}\dheeraj{Yes, you can just mention like you mentioned in Figure 5.}
}
\vspace*{-0.5cm}
\label{acrossBkts}
\end{figure}

\begin{table*}[t]
\small 
    \caption{Performance using multiple multilingual models. First three rows denote performance without using progressive training and the last row denotes the performance when the model with best performance is used with  progressive training. 
    % \dheeraj{If this table is only for Indian language LMs, we don't need to add spanish-english column at all.} \sud{The table doesn't look very convincing if we remove spanish. Please see below.}
    }
    \centering
  \resizebox{\linewidth}{!}{
  
          \begin{tabular}{lccccccccc} 
        %   \hline
        \toprule
          \textbf{Model} & \multicolumn{3}{c}{\textbf{Spanish-English}} & \multicolumn{3}{c}{\textbf{Hindi-English}} & \multicolumn{3}{c}{\textbf{Tamil-English}} \\ 
    \cmidrule(lr){2-4} \cmidrule(lr){5-7} \cmidrule(lr){8-10}
          & Macro-F1 & Micro-F1 & Weighted-F1 & Macro-F1 & Micro-F1 & Weighted-F1 & Macro-F1 & Micro-F1 & Weighted-F1\\ 
          \midrule
        mBERT &  $76.2 \pm 0.6$	 & $76.3 \pm 0.6$	 & $76.3 \pm 0.6$   & $67.1 \pm 0.8$ &	$69.2 \pm 0.3$ &	$68.4 \pm 0.6$ &  $53.3 \pm 0.4 $ &	$78.6 \pm 0.6 $ &	$76.7 \pm 0.2 $\\
        MuRIL	&	- & - & - &		$\mathbf{77.0 \pm 0.4} $ &	$\mathbf{77.7 \pm 0.3} $ &	$\mathbf{77.7 \pm 0.4} $ &	$54.2 \pm 0.2 $ &	$64.2 \pm 0.4 $ &	$68.8 \pm 0.3 $ 	\\ 
        % \hline
        IndicBERT &	- & - & - &			$73.5 \pm 0.5$ &	$74.5 \pm 0.3$ &	$74.3 \pm 0.4$ &	$\mathbf{54.6 \pm 0.1}$ &	$68.0 \pm 0.6$ &	$71.3 \pm 0.4$	\\ 
        % \hline
        Ours + Best &	$\mathbf{77.4 \pm 0.8}$	 & $\mathbf{77.5 \pm 0.8}$	 & $\mathbf{77.5 \pm 0.8}$  &	$\mathbf{77.0 \pm 0.4}$ &	$77.6 \pm 0.4$ &	$77.6 \pm 0.4$ &	$53.1 \pm 0.4$ &	$\mathbf{80.5 \pm 0.2}$ &	$\mathbf{77.5 \pm 0.1}$ \\
        \bottomrule
        % \hline
        \end{tabular}
    }

  \label{tabModels}
\end{table*}

% \begin{table*}[t]
% \small 
%     \caption{Performance using multiple multilingual models. First three rows denote performance without using progressive training and the last row denotes the performance when the model with best performance is used with  progressive training. }
%     \centering
%   \resizebox{0.85\linewidth}{!}{
%           \begin{tabular}{lcccccc} 
%         %   \hline
%         \toprule
%           \textbf{Model} & \multicolumn{3}{c}{\textbf{Hindi-English}} & \multicolumn{3}{c}{\textbf{Tamil-English}} \\ 
%     \cmidrule(lr){2-4} \cmidrule(lr){5-7} 
%           & Macro-F1 & Micro-F1 & Weighted-F1 & Macro-F1 & Micro-F1 & Weighted-F1 \\ 
%           \midrule
%         mBERT &   $67.1 \pm 0.8$ &	$69.2 \pm 0.3$ &	$68.4 \pm 0.6$ &  $53.3 \pm 0.4 $ &	$78.6 \pm 0.6 $ &	$76.7 \pm 0.2 $\\
%         MuRIL	& $\mathbf{77.0 \pm 0.4} $ &	$\mathbf{77.7 \pm 0.3} $ &	$\mathbf{77.7 \pm 0.4} $ &	$54.2 \pm 0.2 $ &	$64.2 \pm 0.4 $ &	$68.8 \pm 0.3 $ 	\\ 
%         % \hline
%         IndicBERT &		$73.5 \pm 0.5$ &	$74.5 \pm 0.3$ &	$74.3 \pm 0.4$ &	$\mathbf{54.6 \pm 0.1}$ &	$68.0 \pm 0.6$ &	$71.3 \pm 0.4$	\\ 
%         % \hline
%         Ours + Best &	$\mathbf{77.0 \pm 0.4}$ &	$77.6 \pm 0.4$ &	$77.6 \pm 0.4$ &	$53.1 \pm 0.4$ &	$\mathbf{80.5 \pm 0.2}$ &	$\mathbf{77.5 \pm 0.1}$ \\
%         \bottomrule
%         % \hline
%         \end{tabular}
%     }

%   \label{tabModels}
% \end{table*}

\begin{table*}[!ht]
\small 
  \caption{Model performance on the three datasets for different number of buckets ($k$) using mBERT.}
  
  \centering
    \resizebox{\linewidth}{!}{
          \begin{tabular}{lccccccccc}
          \toprule
          \textbf{Buckets} & \multicolumn{3}{c}{\textbf{Spanish-English}} & \multicolumn{3}{c}{\textbf{Hindi-English}} &
          \multicolumn{3}{c}{\textbf{Tamil-English}} \\
    \cmidrule(lr){2-4} \cmidrule(lr){5-7} \cmidrule(lr){8-10}
           & Macro-F1 & Micro-F1 & Weighted-F1 & Macro-F1 & Micro-F1 & Weighted-F1 & Macro-F1 & Micro-F1 & Weighted-F1\\ 
        %   \hline
        \midrule
        k=2 & $\mathbf{77.4 \pm 0.8}$	 & $\mathbf{77.5 \pm 0.8}$	 & $\mathbf{77.5 \pm 0.8}$ & $\mathbf{67.8 \pm 0.8}$ &	$\mathbf{69.9 \pm 0.5}$ &	$\mathbf{69.1 \pm 0.7}$ &	
          $\mathbf{53.1 \pm 0.4}$ &	$80.5 \pm 0.2$ &	$\mathbf{77.5 \pm 0.1}$ \\ 
        %   \hline
        k=3 & $76.1\pm 0.7 $ &	$76.2\pm 0.7 $ &	$76.2\pm 0.7 $ &		$67.2\pm 0.6 $ &	$69.4\pm 0.5 $ &	$68.6\pm 0.6 $ &		$53.1\pm 0.2 $ &	$79.9\pm 0.4 $ &	$77.2\pm 0.1 $ \\
        % \hline
        k=4 & $76.6 \pm 1.6$ &	$76.7 \pm 1.7$ &	$76.7 \pm 1.6$ &		$67.6 \pm 0.4$ &	$69.7 \pm 0.3$ &	$68.9 \pm 0.4$ &		$52.9 \pm 0.4$ &	$\mathbf{80.6 \pm 0.7}$ &	$\mathbf{77.5 \pm 0.3}$ \\
        \bottomrule 
        \end{tabular}
    }
\label{tabKK}
\end{table*}

\textbf{Is the fraction of OOD samples same for both the buckets for model $m_{pt}$?} In the first experiment, we consider the model trained on the source dataset and try to find the fraction of OOD samples in both the buckets. Since the first bucket contains more resource-rich language dominated samples, we expect a lesser fraction of samples to be out-of-distribution compared to the second bucket. In Figure \ref{acrossBkts}, we plot the bucketwise OOD for different datasets. We observe that lesser fraction of samples from first bucket are OOD in all the datasets except Spanish-English.
% We can observe this is true for almost all the datasets in the Figure ~\ref{acrossBkts}. 
% \dheeraj{Mention the dataset names on top}
% However, the difference in the out-of-distribution fraction for the buckets is different across different datasets. 
This shows that instances dominated by resource-rich language are less likely to be out-of-distribution for the classifier trained on $S$ compared to instances dominated by low-resource language, thus providing empirical evidence in support of the first part of our hypothesis.  
We also report the zero-shot performance in Appendix \ref{zeroShotApp} i.e. we use the classifier trained on source dataset for inference with no training on target dataset. For all the datasets, we observe zero-shot model performs better on $B_1$ compared to $B_2$, thus further bolstering our hypothesis.

\begin{figure}[!h]
\centering
\includegraphics[width=0.49\textwidth]{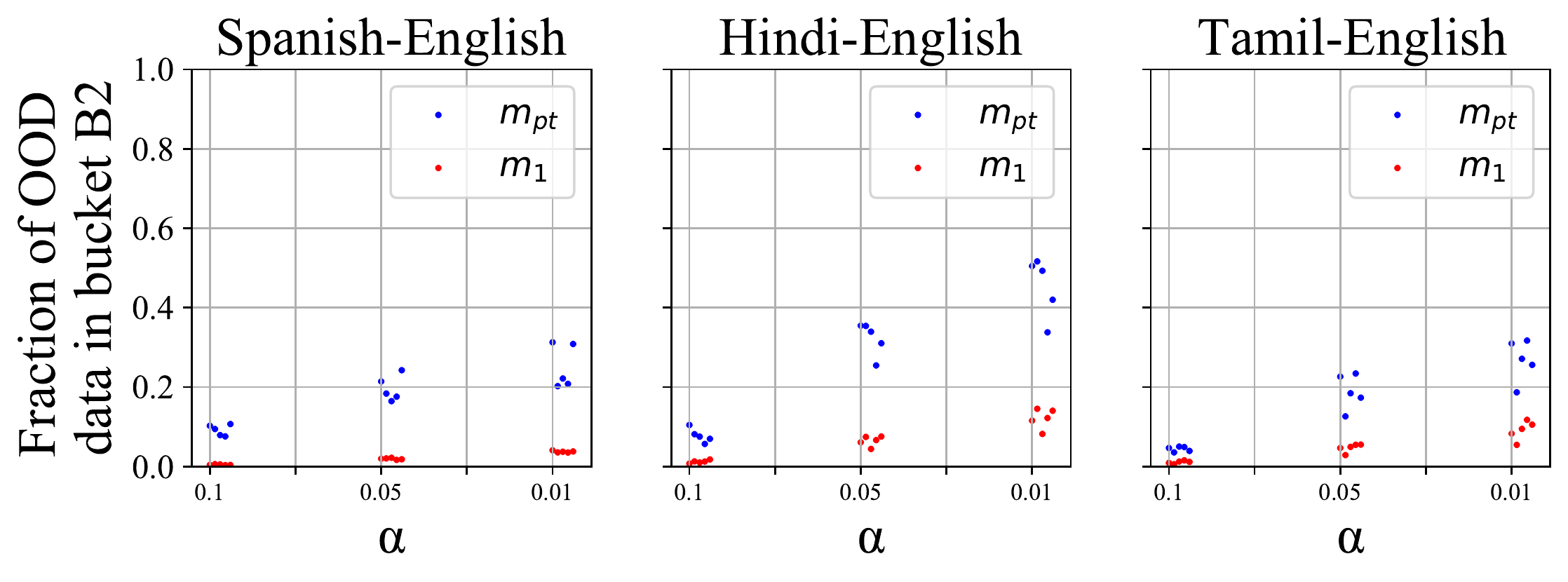} % Reduce the figure size so that it is slightly narrower than the column.
\caption{Figure showing model wise OOD for bucket $B_2$ across multiple datasets. We compare two models, $m_{pt}$ and $m_1$. In all the cases, we observe that samples in $B_2$ are more OOD for model $m_{pt}$ compared to $m_1$ across different values of $\alpha$. Values are reported for 5 runs and points on the same vertical line are from the same run i.e. both $m_{pt}$ and $m_1$ were initialized with same initial weights.
% \dheeraj{Change the colors. May be choose like in Figure 3.}
}
\label{acrossModels}
\vspace*{-0.5cm}
\end{figure}

\textbf{Is there a change in OOD fraction for bucket $B_2$ if we use model $m_1$ instead of model $m_{pt}$?} 

% \sud{Need to Re-write this para}
% \begin{center

In the second experiment, we compare the fraction of OOD data in bucket $B_2$ for the models $m_{pt}$ and $m_1$. 

In Figure \ref{acrossModels}, we observe a lesser fractions of samples in bucket B2 are OOD for model $m_1$ compared to model $m_{pt}$.
% can see across all the datasets that the fraction of the samples in the bucket $B_2$ becomes less out-of-distribution for model $m_1$. 
This is expected since the model $m_1$ has seen samples with low-resource language words while training, thus providing empirical evidence in the support of our proposed training strategy. Although, the samples from $B_2$ would still have noisy labels, we expect them to be more accurate when predicted by $m_1$ than $m_{pt}$.
% \dheeraj{Two "noisy" words in the same sentence is a bit confusing. Can you change this last sentence?}. 

\subsection{Comparison with other multilingual models}
Recently, multiple multilingual transformer models have been proposed.
% research has focused on creating multilingual transformer models focusing on Indian languages. 
We experiment with MuRIL \citep{khanuja-muril} and IndicBERT \citep{kakwani-etal-2020-indicnlpsuite}. 
Firstly, we obtain the performance of three language models: mBERT, MuRIL, and IndicBERT without progressive training on all datasets and we use progressive training on top of the best performing model corresponding to each dataset and verify whether it further improves the performance.
% Firstly we obtain the value using the No-PT method for the three models: mBERT, MuRIL and IndicBERT. Then we take the best performing model and use the progressive training methodology on top of it to see if our method improves the performance. 
The F1 scores are reported in Table \ref{tabModels}. We observe that performance either increases or stays very competitive in all the cases, thus showing our method is capable of improving performance even when used with the best multilingual model for the task. 

\vspace*{-0.2cm}
\subsection{Hyper-parameter sensitivity analysis}
\label{hpSens}
There are two hyper-parameters in our experiments: the number of buckets ($k$) and the ratio of samples selected for self-training ($\delta$). 
We vary $k$ from 2 to 4 to study the effect of the number of buckets on the performance and the F1-scores are reported in Table~\ref{tabKK}.
% We vary the number of buckets from 2 to 4. The values are reported in the table \ref{tabKK}. 
% We observe that the values with k=2 perform either better or competitive with other values. As mentioned earlier, we believe this is because of the number of spikes in the distribution plot of $f_{eng}$ being 1 or 2 across the datasets. In presence of more number of spikes, higher value of $k$ should give better performance.  
Our method is fairly robust to the values of k. For almost all values of k, our method does better than the baselines. As mentioned earlier, the number of spikes in the distribution plot of $f_{eng}$ is 1 or 2 for all the datasets. 
In presence of more number of spikes, higher value of $k$ is recommended. 
% \sud{Check this}
% We believe this is because of small dataset size and in presence of larger datasets, we should have more improvement using a higher value of $k$.\dheeraj{This is different from the reasoning we gave earlier i.e. number of spikes. You need to mention spikes here to be consistent.}
\begin{figure}[h]
\hspace*{-0.2cm}\includegraphics[width=0.50\textwidth]{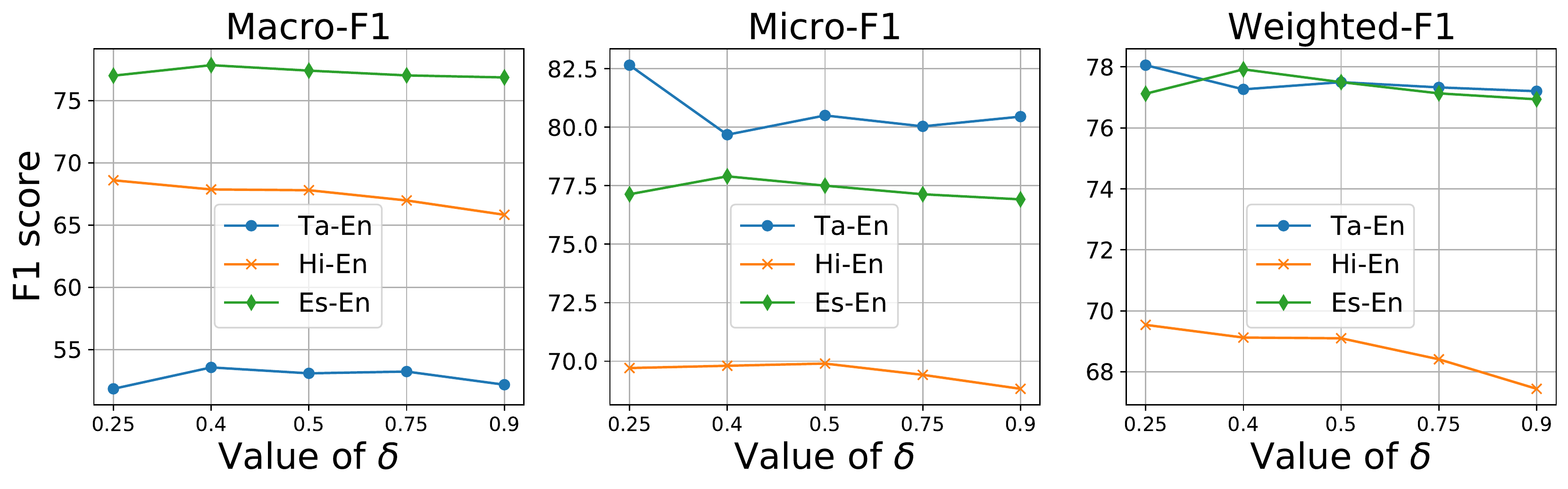} 
\caption{F1 score using different values of parameter $\delta$ for the three datasets (Spanish-English, Hindi-English and Tamil-English). The plots represent macro, micro and weighted F1 score (left-to-right). }
\label{figDelta}
\vspace*{-0.5cm}
\end{figure}
For studying the effect of hyper-parameter $\delta$, we plot macro, micro, and weighted F1 scores across multiple values of $\delta$ in Figure \ref{figDelta}. 
With low $\delta$, there wouldn't be enough sentences for self-training to help whereas with high $\delta$, the samples would be too noisy. Thus, a value in the middle i.e. 0.4-0.6 should be reasonable choice.

\section{Conclusion and Future work}
In this paper, we propose progressive training framework that takes distinction between low-resource and resource-rich language into account while doing zero-shot transfer learning for code-switched sentiment analysis. We show that our framework improves performance across multiple datasets. Further, we also create probing tasks to provide empirical evidence in support of our hypothesis. In future, we want to extend the framework to other tasks like question-answering and natural language inference. 

\section{Limitations}
A key potential limitation of the current framework is that if the number of samples in buckets are very disproportionate, the progressive learning might not result in significant improvement.
% depending on the size of $S$ and the capacity of the model, the model might forget information relevant for the low-resource language. In future, we would like to perform a systematic study of the dependency on size of $S$.
% \dheeraj{can you explain this a bit more?} \sud{Not sure}
% Thus, the original training objective should be modified to account for this. Secondly, there should be a more systematic study on how large does $S$ need to be for this phenomena to occur.

% We also want to explore more efficient ways to transfer knowledge from resource-rich language data to code-switched data. 
% The second is to use our framework for improving the performance when doing zero-shot transfer learning for sentiment analysis across datasets in two different languages using synthetic code-switched text. For example, when doing zero-shot transfer learning from a labelled English dataset to an unlabelled Hindi datatset, we can create synthetic code-switched text and use our progressive training framework.
% \showthe\font
% \expandafter\show\the\font

\section{Ethical consideration}
 This paper proposes a progressive training framework to transfer knowledge from resource-rich language data to low-resource code-switched data. We work on sentiment classification task which is a standard NLP problem. Based on our experiments, we don't see any major ethical concerns with our work.
%  We don't see any significant harmful effect of our work. Our framework aims at solving the sentiment classification task which is a very standard task. We focus on improving performance for code-switching scenario involving low-resource languages. Our work should help in getting better performances for languages that have not been researched extensively. One minor issue might be the biases introduced during transfer learning but we didn't see any such issue in our experiments.
% %  Our work should help in solving tasks related to code-switching with better performance.

\section{Acknowledgements}
We thank Zihan Wang for valuable discussions and comments. Our work is sponsored in part by National Science Foundation Convergence Accelerator under award OIA-2040727 as well as generous gifts from Google, Adobe, and Teradata. Any opinions, findings, and conclusions or recommendations expressed herein are those of the authors and should not be interpreted as necessarily representing the views, either expressed or implied, of the U.S. Government. The U.S. Government is authorized to reproduce and distribute reprints for government purposes not withstanding any copyright annotation hereon.

\bibliography{anthology,custom}
\bibliographystyle{acl_natbib}

\appendix
\clearpage
\newpage

\section{Appendix}

% \subsection{Additional dataset details} \label{newL}
% Most of the sentences in the datasets are written in the Roman script. The words in Hindi and Tamil are converted into the Devanagari script. We use the processed dataset provided by GLUECoS for Hindi-English and Spanish-English datasets. The processed version of Hindi-English dataset has the Hindi words written in Devanagari script. Since we deal with low-resource languages for which tools might not be well developed, we use heuristics to detect the words in English. For Spanish-English  and Tamil-English datasets, we use the Brown corpus from NLTK \footnote{https://www.nltk.org/} to detect the English words in the sentence. Words that are not present in the corpus are considered of another language. For the Tamil-English dataset, we use the AI4Bharat Transliteration python library \footnote{https://pypi.org/project/ai4bharat-transliteration/} to get the transliterations. 

\subsection{Additional hyper-parameter details} \label{hyperP}
% For pretraining with supervised English dataset, we use 4 epochs and choose the best weights using the validation set. For training, we use 4 epochs in all the iterations and use the weights obtained after fourth epoch. For final evaluation, we use the model obtained from the last iteration. 
The batch size is 64, sequence length is 128 and learning rate is 5e-5. These hyperparameters are same for all the models used during pre-training and training. Every iteration takes approximately $\sim$1-2 seconds and $\sim$12 GB of memory on a GPU. 

% Most of the sentences in the datasets are written in the Roman script. The words in Hindi and Tamil are converted into the Devanagari script. We use the processed dataset provided by GLUECoS for Hindi-English and Spanish-English dataset. For Hindi-English dataset, GLUECoS used the LID-tool \footnote{https://github.com/microsoft/LID-tool} for language identification.
% % For the  Hindi-English dataset, the LID-tool \footnote{https://github.com/microsoft/LID-tool} is used for language identification. 
% To do language identification in Spanish-English  and Tamil-English datasets, we use the Brown corpus from NLTK \footnote{https://www.nltk.org/} to filter out the English words in the sentence. Words that are not present in the corpus are considered of another language. 
% For the Tamil-English dataset, we use the AI4Bharat Transliteration python library \footnote{https://pypi.org/project/ai4bharat-transliteration/} to get the transliterations. 

\subsection{Statistics related to the buckets} \label{bktStats}
\begin{figure}[H]
\includegraphics[width=0.46\textwidth]{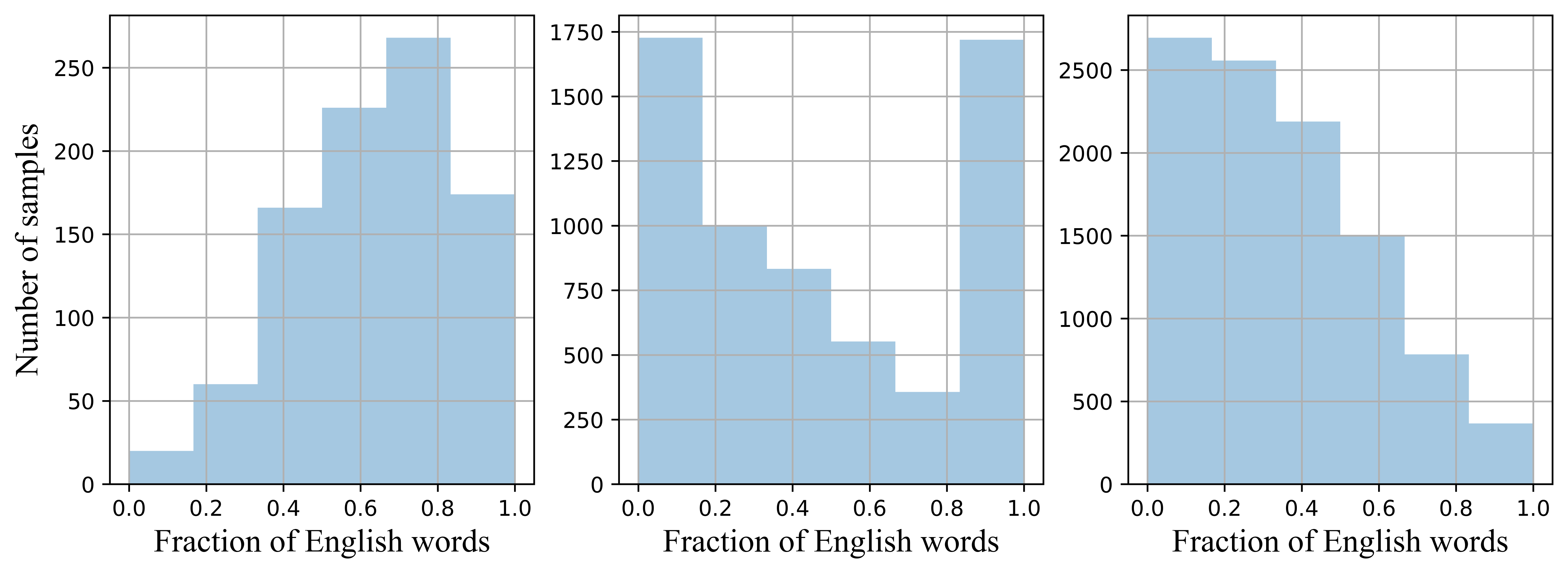} \hfill
\caption{Distribution of $f_{eng}(x)$ vs number of samples for the Spanish-English, Hindi-English and Tamil-English datasets (left-to-right). For Hindi-English, we can observe two spikes in the graph showing some samples are heavily dominated by English and some samples are heavily dominated by Hindi. For the other two datasets, we observe the progression to be more gradual.}
\label{distPlotImg}
\end{figure}

We report the average value and standard deviation of the $f_{eng}$ across the buckets in Table \ref{fracEng}. We report the number of instances selected for self-training across the buckets in Table \ref{cntEng}. We plot the distribution of $f_{eng}$ in Figure \ref{distPlotImg}.

\begin{table}[!h]
    \centering
    \caption{Average value and standard deviation of $f_{eng}$ for both the buckets $B_1$ and $B_2$.}
      \resizebox{\linewidth}{!}{

\begin{tabular}{|c|c|c|}
\hline
\textbf{Dataset} & \textbf{$B_1$} & \textbf{$B_2$} \\ \hline
Spanish-English & $0.79 \pm 0.08$ & $0.44 \pm 0.14$    \\ \hline
Hindi-English & $0.79 \pm 0.21$ & $0.14 \pm 0.13$   \\ \hline
Tamil-English & $0.51 \pm 0.16$ & $0.13 \pm 0.09$  \\ \hline
\end{tabular}
}
    \label{fracEng}
\end{table}

\begin{table}[!h]
    \centering
    \caption{Number of samples selected from each bucket B1 and B2 for progressive training.}
        %   \resizebox{\linewidth}{!}{

% \small
\begin{tabular}{|c|c|c|}
\hline
\textbf{Dataset} & \textbf{$B_1$} & \textbf{$B_2$} \\ \hline
Spanish-English & 265 & 191   \\ \hline
Hindi-English & 2047 & 1048  \\ \hline
Tamil-English & 2877 & 2169  \\ \hline
\end{tabular}
% }
    \label{cntEng}
\end{table}

% \vfill\null
% \columnbreak
% \vfill\null
% \vfill\break

% \subsection{Distribution plot of $f_{eng}$ words} \label{fracDiag}

% \dheeraj{Merge appendix sections A.3 and A.4 since both are about buckets}

\subsection{Results of the zero-shot model on buckets} \label{zeroShotApp}
We report the results of the zero-shot model on both the buckets in Table \ref{zeroShotRes}. As expected, in all the cases the model performs better on $B_1$ compared to $B_2$. 

\begin{table}[!ht]
    \centering
        \caption{Bucketwise results of the zero-shot model. Macro, Micro and Wtd refer to Macro-F1, Micro-F1 and Weighted-F1 scores respectively.}

      \resizebox{\linewidth}{!}{

    \begin{tabular}{ccccccc}
        \toprule
           & \multicolumn{3}{c}{\textbf{$B_1$}} & \multicolumn{3}{c}{\textbf{$B_2$}} \\
    \cmidrule(lr){2-4} \cmidrule(lr){5-7} 
    \textbf{Dataset} & \textbf{Macro} & \textbf{Micro} & \textbf{Wtd} & \textbf{Macro} & \textbf{Micro} & \textbf{Wtd} \\ \midrule
        Spanish-English & 79.2 & 79.6 & 79.7 & 71.8 & 73.3 & 73.8\\ 
        Hindi-English  & 73.9 & 77.6 & 76.9 & 56.0 & 58.0 & 56.2 \\ 
        Tamil-English  &54.6 & 83.4 & 81.6 & 50.6 & 77.0 & 72.8\\
        \bottomrule
        \end{tabular}
        }
    \label{zeroShotRes}
\end{table}

\subsection{Statistical Significance Results}
\label{statSigApp}
\begin{table}[!ht]
    \centering
        \caption{We perform paired t-test between our method and baselines. The p-value obtained by performing the test between our methods and baselines for all three datasets is reported in the table.}

      \resizebox{\linewidth}{!}{

    \begin{tabular}{|c|c|c|c|c|}\hline
        \textbf{Dataset} & \textbf{No-PT} & \textbf{DCE} & \textbf{\textit{- Source}} & \textbf{\textit{- Ratio}} \\ \hline
        Spanish-English      &  $5.11e^{-13}$ & $2.09e^{-6}$ & $8.28e^{-4}$ &  $8.82e^{-8}$\\ \hline
        Hindi-English  &  $2.76e^{-6}$ & $4.37e^{-43}$ & $3.49e^{-4}$ & $6.85e^{-12}$\\ \hline
        % Ma-En FIRE &  $3.09e^{-5}$  & $1.50e^{-5}$ & $2.2e^{-4}$\\ \hline
        Tamil-English  &  $1.87e^{-41}$ & $ 3.72e^{-16}$ & $3.71e^{-3}$ & $3.40e^{-7}$\\
        \hline
        \end{tabular}
        }
    \label{significanceTests}
\end{table}

\newpage
\subsection{Qualitative analysis} \label{qualAna}
As discussed previously, on the low-resource language dominated bucket, our model is correct more often than the \textbf{\textit{No-PT}} baseline. We focus on samples from bucket $B_2$ for qualitative analysis. For the sample, \textit{"fixing me saja hone ka gift"}, the Hindi word \textit{"saja"} refers to punishment which is negative in sentiment whereas the word \textit{"gift"} is positive in sentiment. Thus, the contextual information in the Hindi combined with that of the English is necessary to make correct prediction.  For the sample \textit{"Mera bharat mahan, padhega India tabhi  badhega India"}, the model has to identify Hindi words "mahan" \& "badhega" to make the correct predictions. 

We also do the qualitative analysis by looking at predictions of samples between successive iterations. In Table~\ref{qam01}, we randomly choose samples which are predicted incorrectly by model $m_{pt}$ but are predicted correctly by model $m_1$. Out of 8 samples, 6 samples had sentiment specifically present in the Hindi words. In Table \ref{qam12}, we randomly choose samples which are predicted incorrectly by model $m_1$ but are predicted correctly by model $m_2$. Out of 8 samples, 4 samples had sentiment specifically present in the Hindi words and 2 samples required understanding both the Hindi and English words simultaneously. The blue highlighted words are relevant to determining the sentiment of the sentence. 
% \sud{No of samples per bucket has to be same?}

% \sud{Effect of Lid not found yet}

% \sud{Figures change / reorder}
% \dheeraj{Mention what the blue highlight in table 8 and 9 eans}

% \begin{table}[!ht]
%     \centering
%     \caption{Instances where model $m_{pt}$ was incorrect but model $m_1$ learnt the correct label.}
% \begin{tabular}{p{0.15\linewidth}  p{0.6\linewidth}  p{0.15\linewidth}}
% \hline 
% \textbf{Bucket} & \textbf{Text} & \textbf{Label} \\ \hline
% 0 & yes we teachers really need \textcolor{blue}{sar dard ki goli} aftr taking class & negative\\ \hline
% 0 & they are promising moon right now to get the cm post . . . waade aise hone chahiye jo janta k welfare k liye ho . . . \textcolor{blue}{free wahi baantna chahta hai jo} \textcolor{blue}{desperate ho kisi tarah ek baar} \textcolor{blue}{ bas kursi mil jaye} . & negative \\ \hline
% 0 & oooo ! grandfather bas ab \textcolor{blue}{nahi kitna natak karoge}	& negative\\ \hline
% 0 & i \textcolor{blue}{agree} with this cartoon . \textcolor{blue}{bahut achcha doston} . & positive \\ \hline
% 0 & seriously \textcolor{blue}{maza boht} ata tha , , , mere pass \textcolor{blue}{mast collection } hota tha & positive \\ \hline
% 1 & school mein milne wale laddoo kitni \textcolor{blue}{khushi} dete they :d & positive\\ \hline
% 1 & guddu itni paas se tv dekhega toh \textcolor{blue}{aankhe button ho jayengi} ! & negative \\ \hline
% 1 & arun lal ki commentary yaad aa gayi . . usse zyada \textcolor{blue}{manhoos} koi nahi . & negative\\ \hline
% \end{tabular}
%     \label{qam01}
% \end{table}

\newpage

\begin{table}[!ht]
\centering
    \caption{Instances where model $m_{pt}$ was incorrect but model $m_1$ learnt the correct label.}

\begin{tabular}
{p{0.15\linewidth}  p{0.6\linewidth}  p{0.15\linewidth}}
\hline 
\textbf{Bucket} & \textbf{Text} & \textbf{Label} \\ \hline
0 & yes we teachers really need \textcolor{blue}{sar dard ki goli} aftr taking class & negative\\ \hline
0 & they are promising moon right now to get the cm post . . . waade aise hone chahiye jo janta k welfare k liye ho . . . \textcolor{blue}{free wahi baantna chahta hai jo} \textcolor{blue}{desperate ho kisi tarah ek baar} \textcolor{blue}{ bas kursi mil jaye} . & negative \\ \hline
0 & oooo ! grandfather bas ab \textcolor{blue}{nahi kitna natak karoge}	& negative\\ \hline
0 & i \textcolor{blue}{agree} with this cartoon . \textcolor{blue}{bahut achcha doston} . & positive \\ \hline
0 & seriously \textcolor{blue}{maza boht} ata tha , , , mere pass \textcolor{blue}{mast collection } hota tha & positive \\ \hline
1 & school mein milne wale laddoo kitni \textcolor{blue}{khushi} dete they :d & positive\\ \hline
1 & guddu itni paas se tv dekhega toh \textcolor{blue}{aankhe button ho jayengi} ! & negative \\ \hline
1 & arun lal ki commentary yaad aa gayi . . usse zyada \textcolor{blue}{manhoos} koi nahi . & negative\\ \hline
\end{tabular}
    \label{qam01}
    
\end{table}

\begin{table}[!htb]
    \centering
    \caption{Instances where model $m_1$ was incorrect but model $m_2$ learnt the correct label.}
\begin{tabular}{p{0.15\linewidth}  p{0.6\linewidth}  p{0.15\linewidth}}
\hline 
\textbf{Bucket} & \textbf{Text} & \textbf{Label} \\ \hline
0 & \textcolor{blue}{Osm} party salman bhai apka gift	& positive\\ \hline
0 & mat dekho brothers and sisters , \textcolor{blue}{dekha nahi jayega}	 & negative \\ \hline
0 & Ap late ho ap ne apne \textcolor{blue}{cometment pori nai ki}	& negative\\ \hline
0 & looks like \textcolor{blue}{'kaho na pyaar hai' phase ended} for modi \#aapsweep \#aapkidilli \#delhidecides & negative \\ \hline
1 & comment krne se jyada sabke padhne me \textcolor{blue}{maja} aata h	& positive\\ \hline
1 & guddu tumhara hi school aisa hoga yaar . mera school n mere teacher to \textcolor{blue}{ache} h &	positive \\ \hline
1 & sir kya msg krtte kartte \textcolor{blue}{mar jau} & negative\\ \hline
1 & Hum \textcolor{blue}{bhuke mur} rahe hai sir & negative\\ \hline
\end{tabular}
    \label{qam12}

\end{table}

% \section{Example Appendix}
% \label{sec:appendix}

% This is an appendix.

\end{document}